\documentclass{article}
\usepackage{spconf}
\usepackage{cite}
\usepackage{amsmath,amssymb,amsfonts}
\usepackage{algorithmic}
\usepackage{graphicx}
\usepackage{textcomp}
\usepackage{xcolor}
\usepackage{amsmath, amsfonts, times, epsfig, amssymb,  blindtext, float, subcaption, multicol, array, multirow, graphicx, etoolbox,  accents, tabularx, tabu, comment, soul}
\usepackage{color}
\usepackage{booktabs}
\usepackage{colortbl}% rowcolor
\makeatletter
\let\NAT@parse\undefined
\makeatother
\usepackage[backref=page,colorlinks=true]{hyperref}
\usepackage{cite}

\begin{document}
\ninept

\newcommand\blfootnote[1]{%
  \begingroup
  \renewcommand\thefootnote{}\footnote{#1}%
  \addtocounter{footnote}{-1}%
  \endgroup
}

\def\etal{\emph{et al}.}
\def\ie{\emph{i.e.,}}
\def\eg{\emph{e.g.,}}
\definecolor{tablegray}{gray}{.9}

% \title{\vspace{-2cm}Improving 3D Pose Estimation for Sign Language\vspace{-0.25cm}}
\title{Improving 3D Pose Estimation for Sign Language}
% \title{Improving 3D Pose Estimation for Sign Language\vspace{-0.25cm}}

% \author{\IEEEauthorblockN{Maksym Ivashechkin}
% \IEEEauthorblockA{
% \textit{Unversity of Surrey, UK\\
% m.ivashechkin@surrey.ac.uk}}
% \and
% \IEEEauthorblockN{Oscar Mendez}
% \IEEEauthorblockA{
% \textit{Unversity of Surrey, UK\\
% o.mendez@surrey.ac.uk}}
% \and
% \IEEEauthorblockN{Richard Bowden}
% \IEEEauthorblockA{
% \textit{Unversity of Surrey, UK\\
% r.bowden@surrey.ac.uk}
% }}\vspace{-0.25cm}

% \name{\vspace{-0.3cm}Maksym Ivashechkin, Oscar Mendez, Richard Bowden}
\name{Maksym Ivashechkin, Oscar Mendez, Richard Bowden}
\address{University of Surrey\\
Centre for Vision, Speech and Signal Processing (CVSSP)\\
 Stag Hill, University Campus, Guildford GU2 7XH, UK}

% \twoauthors
 % {Maksym Ivashechkin, Oscar Mendez, Richard Bowden}
	% {\textit{University of Surrey, UK, CVSSP, GU2 7XH}
 % {C. Author-three, D. Author-four\sthanks{The fourth author performed the work
	% while at ...}}
	% {School C-D\\
	% Department C-D\\
	% Address C-D}
 % {A. Author-one, B. Author-two\sthanks{Thanks to XYZ agency for funding.}}
	% {School A-B\\
	% Department A-B\\
	% Address A-B}
 % {C. Author-three, D. Author-four\sthanks{The fourth author performed the work
	% while at ...}}
	% {School C-D\\
	% Department C-D\\
	% Address C-D}

\maketitle

\begin{abstract}
This work addresses 3D human pose reconstruction in single images.
We present a method that combines Forward Kinematics (FK) with neural networks to ensure a fast and valid prediction of 3D pose.
Pose is represented as a hierarchical tree/graph with nodes corresponding to human joints that model their physical limits.
Given a 2D detection of keypoints in the image, we lift the skeleton to 3D using neural networks to predict both the joint rotations and bone lengths. These predictions are then combined with skeletal constraints using an FK layer implemented as a network layer in PyTorch. The result is a fast and accurate approach to the estimation of 3D skeletal pose.  
Through quantitative and qualitative evaluation, we demonstrate the method is significantly more accurate than MediaPipe in terms of both per joint positional error and visual appearance. Furthermore, we demonstrate generalization over different datasets and sign languages. The implementation in PyTorch runs at between 100-200 milliseconds per image (including CNN detection) using CPU only.
%The primary contribution of our work is application on a sign language data where hands and body interactions are very important.

% The approach is estimation of Euler angles by  to parametrize and constrain an individual human's joint.
% Despite Euler angles rotation representation is being misjudged 
% We show that a simple multiple layer perceptron is enough for accurate prediction of a 3D pose.
% The method 
\end{abstract}


% \vspace{-0.5cm}
\section{Introduction}
% \vspace{-0.24cm}
Human pose estimation is an active and challenging field of research and recent years have seen significant progress in deep learning approaches to the estimation of 2D human key-points in images~\cite{mmpose2020}. However, the breadth of potential applications, associated variability in appearance and the complexity of human motion make this an extremely challenging task. Lifting 2D to 3D (or estimating human pose directly in 3D) has inherent ambiguities which must be overcome using subtle visual cues. This often involves reasoning and/or complex statistical priors or 3D meshes~\cite{SMPL:2015}. 

% Recent work has shown that pose regression is strongly influenced by the domain in which the network is trained [CITE]....

In this work, we demonstrate that domain-specific training of pose regression outperforms existing general solutions, and we can trade generality for accuracy by focusing on the application. %<Something about amounts of training data?>. 
Specifically, we focus on sign language. However, even in the more general case, our simple approach can outperform the \emph{state-of-the-art}.

\begin{figure}
    \centering
    \includegraphics[width=0.75\linewidth]{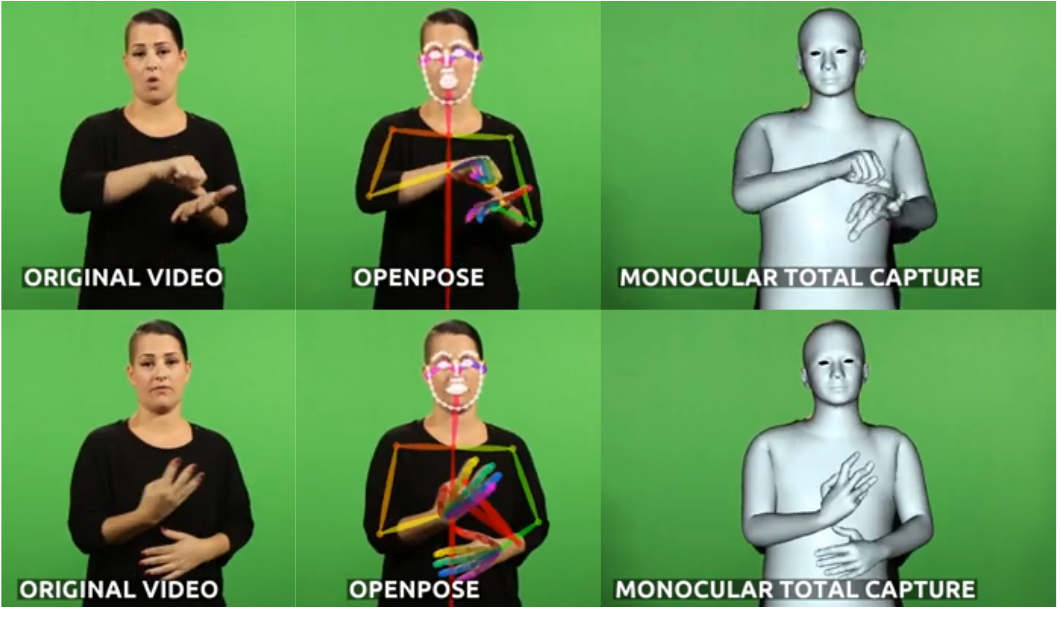}
    \caption{Figure showing OpenPose failure for hand-to-hand interaction and associated incorrectly reconstructed mesh for Monocular Total Capture.
    % \vspace{-0.70cm}
    }
    \label{fig:failure_examples}
\end{figure}

Human pose estimation is an appealing first stage to support sign language recognition as a human skeleton provides natural invariance to a person, clothing, and background. 
However, sign language has many of its own challenges that lead to common failures for generic pose estimation techniques. 
During sign, the hands move quickly resulting in motion blur which leads to keypoint detection failure for frame-based pose estimation techniques.
Furthermore, sign %by its very definition, 
involves extensive hand-to-hand and hand-to-face interactions. 
These types of complex hand-to-body interactions are a common point of failure for pose estimation techniques. 
Fig.~\ref{fig:failure_examples} gives two examples of OpenPose~\cite{openpose} and Monocular Total Capture~\cite{xiang2019monocular} where such hand-to-hand and hand-to-face interactions lead to failures in estimation,~{\eg} ambiguous hand shape, distortion, or position.

\begin{figure*}
    % \vspace{-0.6cm}
    \centering
    \includegraphics[width=0.75\linewidth]{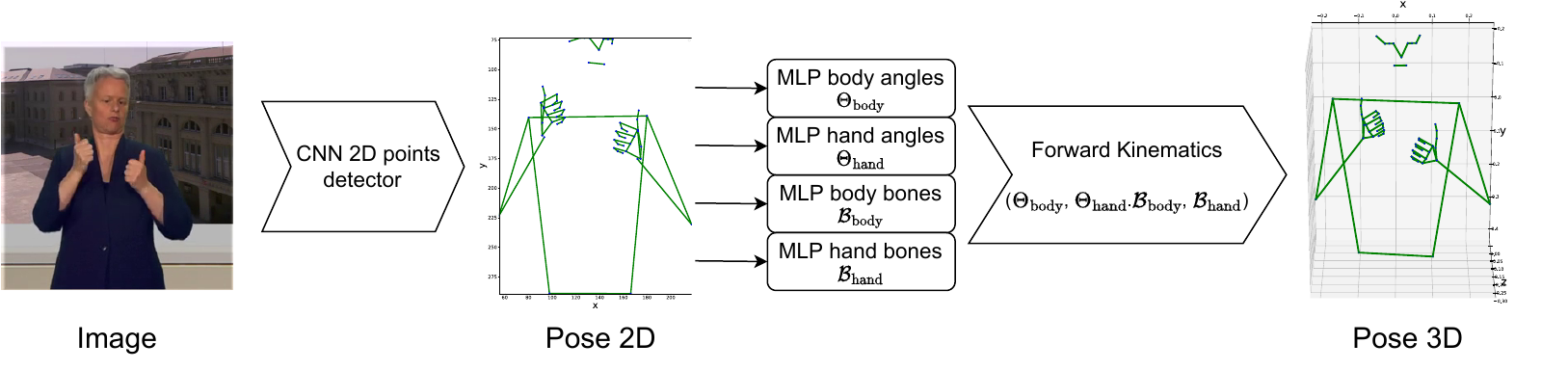}
    \caption{The pipeline for pose reconstruction from a single image. The input image undergoes a 2D detection ({\eg} HRNet~\cite{hrnet} or MediaPipe). Separate networks then generate rotations and bone lengths for the body and hands, and a custom FK layer in PyTorch combines this information to produce a 3D pose. Image source is DSGS EASIER dataset~\cite{kopf_maria_2021_9561}.
    % \vspace{-0.5cm}
    }
    \label{fig:overview}
\end{figure*}

% SUMMARISE THE NOVELTIES OF THE PAPER THEN GIVE AN OVERVIEW OF THE REMAINING SECTIONS
This paper proposes a pose reconstruction pipeline that given a single image uses neural networks to generate human joint orientations and bone-length predictions which are fused with a kinematics model.
% In the next section, background work related to human pose estimation is discussed.
% Section~\ref{section:pose_reconstruction} describes the methodology for human 3D reconstruction.
% Finally, in the experiments, we report accuracy of the presented method and a visual comparison over datasets.

% \vspace{-0.24cm}
\section{Background}
% \vspace{-0.24cm}
There are various techniques available for 3D pose estimation from a single image, which can be broadly categorized into two groups. The first group involves the direct regression of 3D points from the image using either 2D pose key-points or features obtained from an image processing model, such as a convolutional neural network (CNN). The second group of methods utilizes forward kinematics (FK) in combination with the uplift from 2D.

% 3D Pose estimation methods are able to reconstruct a pose from a single image by predicting 2D joints first and then lifting them to 3D; or using the features of the detector ({\eg} convolutional neural networks) to directly predict the 3D pose.
% % Many algorithms work with only 2D landmark points to estimate 3D pose.
% % Commonly used methods either regress pose points by minimizing error in 3D, or find joint angles and generate 3D pose by FK.
% Other algorithms work directly with the 2D points.
% The different approaches can be divided into two groups: (a) regression of 3D points directly from 2D, or (b) combined lifting from 2D with FK.

One design choice is whether to predict joint orientation via the networks rather than position. In QuaterNet~\cite{pavllo2018quaternet} the authors predict 3D rotation via a quaternion representation.
Zhou~{\etal}~\cite{Zhou2019OnTC} show that rotations in 3D have a continuous representation in at least five-dimensional space. 
In~\cite{levinson20neurips}, Levinson~{\etal} demonstrate that a 9D rotation matrix representation provides the best performance in model training compared to other representations. 
%5D, 6D, quaternion, Euler angle, and other representations.
In~\cite{pavllo2018quaternet, Zhou2019OnTC,levinson20neurips} FK is applied to generate new poses after model prediction of rotations given a 3D pose input.

Human 3D pose estimation from a single image via Inverse Kinematics (IK) with an unscented Kalman filter was presented in~\cite{seo_ik}, which shows better results than numerical IK reconstruction and faster convergence. %....
Li~{\etal}~\cite{li2020hybrik}~exploit neural networks to predict 3D pose, then refine it using IK by running a twist-and-swing decomposition~\cite{Baerlocher2001} to estimate the rotations of body joints.
A single image approach was presented in~\cite{denis_lifting} that predicts features and 2D joints from an image, lifts them to 3D, and projects them back to the image to refine the estimate.
Yang~{\etal}~\cite{YANG2022108439} describe pose estimation from images combining 3D regression and pose prediction using Lie algebra with FK.

Pose reconstruction using a feed-forward network was demonstrated by Zhao~{\etal}~\cite{zhao_recover_3d}. In the context of sign, pose estimation using quaternion prediction from 2D followed by FK was described by Krishna~{\etal}~\cite{Krishna_2021_ICCV}.
In Elepose~\cite{elepose}, Wandt~{\etal} propose an unsupervised approach to 3D pose estimation (using only 2D poses input) by selecting the pose with the maximum likelihood over random 2D projections, where the likelihood is found via normalizing flows on 2D poses.

% In this paper, we present a human 3D pose reconstruction approach from a single image. % pose key-points applicable in a sign-language domain.
% We show that the combination of linear networks for predicting angles and FK is able to generate valid and accurate 3D reconstructions.
% Evaluation on different human pose datasets demonstrate that the proposed method outperforms MediaPipe, and it is comparable to other methods.

% \subsection{Related Work}

% \subsection{Contributions}
% \begin{enumerate}
%     \item We provide IK solver that can be able to fit either 3D pose and 2D pose with given calibration. This also allows to generate new data and re-train the proposed model.

% \end{enumerate}

%%%%%%%%%%%%%%%%%%%%%%%%%%%%%%%%%%%%%%%%%%%%%%%%%%%%%%%%%%%%%%%%
% \vspace{-0.9cm}
\section{Pose reconstruction}
% \vspace{-0.4cm}
\label{section:pose_reconstruction}

Our %pipeline to 3D reconstruction of human pose consists of multiple steps. The 
network structure is shown in~Fig.~\ref{fig:overview}.
We use a CNN backbone to extract 2D key-points from an image, followed by four multi-layer perceptrons (MLPs) that generate joint Euler angles and bone lengths.
Finally, a human pose is obtained via FK using the output angles and limb lengths.

A fully-connected linear network (MLP) is used to predict angles of human joints, and the FK propagates joint position and rotations from the root.
% A fully-connected linear network is used to predict one global rotation for the root joint,~{\ie} from where FK propagates joint position and angles.
The rotation is represented via Euler angles. %, despite authors of~\cite{pavllo2018quaternet, Zhou2019OnTC, learning_3d_skeleton} showing that other representations work better for their tasks.
We enumerate several reasons to justify this.
% Although, some papers suggest using other representations for the rotation matrix such as quaternions in~\cite{pavllo2018quaternet}, high dimensional (5D or 6D)~\cite{Zhou2019OnTC}, or nine dimensional using an SVD decomposition~\cite{learning_3d_skeleton}, the proposed network predicts Euler angles for several reasons.
Firstly, it is easier to represent a rotation matrix that has fewer than 3 degrees of freedom (DoF).
This is needed because the human pose consists of many joints with only one or two DoF such as the elbows or fingers. 
Secondly, exploiting Euler angles helps to enforce additional limits and constraints for the rotation of human joints.
For instance, the head can move around 90$^{\circ}$ degrees for roll, pitch, and yaw angles.
% \vspace{-0.6cm}
\subsection{Angular network}
% \vspace{-0.3cm}
The structure of the angular MLP includes several fully-connected layers ({\eg} 128-512 hidden units per layer) followed by a ReLU activation except for the last layer where a sigmoid function is used. 
The network's inputs are 2D pose points flattened into a single vector. 
% , and the outputs are Euler joint angles.
The sigmoid layer normalizes each output angle $\Tilde{x} \in [0, 1]$ range, and the angular constraint is enforced by multiplying the angle's limits,~{\ie} $x = \Tilde{x} \cdot (x_{\max} - x_{\min}) + x_{\min}$.
This step not only guarantees the network outputs valid angles but also solves the problem of non-uniqueness and discontinuity ({\eg} 0 and $2\pi$) of the Euler angle representation%. As mentioned in~\cite{Zhou2019OnTC}
, since angles will always be lower than 180$^\circ$ corresponding to human limits.
The root joint rotation has 3 DoF in the general case and remains unconstrained,~{\ie} it has range $[-\pi; \pi]$ for roll, pitch, and yaw.
This enables a human to rotate, perform a handstand, and lie on their side.
%However, in the context of sign language, we can remove some rotational movement from the root joint if the assumption is, for example, that a person sits upright and reasonably still.

\subsection{Limb lengths network}
The bone length network has similar but smaller architecture than the angular network, because its role is less complex. %/, and even simple regression gives satisfactory prediction.
% The MLP for bone lengths can have a lighter architecture than the angular MLP, because limb lengths are not relevant in context of sign domain and could be guessed with statistical approaches.
We use an MLP with two layers in the experiments (see~\ref{section:experiments}).
%However, sigmoid normalization is removed as it gave worse performance in experiments.
The only constraint enforced for limb length is it being greater than zero, but sigmoid normalization can also be used to constrain bone lengths in the desired range.

% \vspace{-0.2cm}
\subsection{Forward kinematics}
% \vspace{-0.2cm}
The output of the angular network and the MLP for bone length are then combined in an FK layer to produce a 3D pose.
The human pose is represented as a hierarchical tree structure where a root node is the root joint,~{\eg} center of the hips.
The position and rotation of the root joint indicate the origin of a pose and its orientation in the camera frame.

The FK is performed by traversing a tree $\mathcal{T}$ with a breadth-first search algorithm~\cite{Bundy1984}.
The computation of each joint position $\mathbf{p}_i$ and its rotation $\mathbf{R}_i, i>0$ (except for the root's index 0) is performed as 
$%\begin{equation}
    \mathbf{R}_i = \mathbf{R}_j \mathbf{R}^\prime_i%, \:\:\:\:\:\forall (i, j) \in E(T), \text{$j$ is parent of $i$}
$ %\end{equation}
and 
$%\begin{equation}
    \mathbf{p}_i = \mathbf{R}_i\, \mathbf{o}_i + \mathbf{p}_{j}%, \:\:\:\:\:\forall (i, j) \in E(T), \text{$j$ is parent of $i$}
$ %\end{equation}
for each edge $(i, j)$ of the graph $\mathcal{T}$, where $j$ is a parent of $i$; $\mathbf{o}_i$ denotes the offset constructed from a bone length of a joint $i$, and $\mathbf{R}^\prime_i$ is the relative rotation of $0 \le k \le 3$ DoF computed from the predicted angles of the MLP.
The rotation $\mathbf{R}^\prime_i$ with zero DoF equals an identity matrix.
% ,~${\ie}$ the root joint does not have orientation.
For instance, all points on a head (eyes, ears, etc) share only one rotation since they do not move independently. Hence, only one node has a full rotation describing the movement of the whole head while other points inherit this rotation as part of the FK.
The orientation of the root joint equals to the relative rotation predicted by MLP since the root has no parent: $\mathbf{R}_0 = \mathbf{R}^\prime_0$, and root location $\mathbf{p}_0$ (pose origin) defines translation for all other points.%\vspace{-0.05cm}

% \vspace{-0.3cm}
\subsection{Objective function}
% \vspace{-0.2cm}
The angular MLP can be supervised by a single or combination of objective functions:
\begin{enumerate}
    \item absolute difference w.r.t. the ground truth (GT) angles.
    \item Euclidean distance between generated 3D and GT poses.
    \item Euclidean distance between projected 3D and GT image points.
\end{enumerate}
% (1) absolute difference w.r.t. the ground truth (GT) angles, (2) Euclidean distance between generated 3D and GT poses, (3) Euclidean distance between projected 3D and GT image points.

Denote $\mathbf{X}_W \in \mathbb{R}^{3\times N}$ to be a pose of $N$ points in columns in the world frame, and $\mathbf{P} = \mathbf{K} \,[\mathbf{R}\:\: \mathbf{t}]$ is a perspective projection matrix, where $\mathbf{K}$ is the intrinsic camera matrix, and $(\mathbf{R}, \mathbf{t})$ are extrinsic parameters.
The pose $\mathbf{X}_W$ can be converted to pose $\mathbf{X}_C = \mathbf{R} \mathbf{X}_W + \mathbf{t}$ in the camera frame, and its projection onto image plane is $\mathbf{X}_I \sim \mathbf{K} \mathbf{X}_C$.
Each of these objective functions has advantages and disadvantages.
The first loss, when applied to the predicted and the ground truth angles, provides faster training since FK is avoided.
However, the MLP does not explicitly learn anything about the 3D point position, and errors on the root angles imply significantly higher errors on 3D points.
% The MLP predicts joint angles that can be used directly to compute the loss against the ground truth angles.
% The positive side is fast training since FK is avoided, however, the MLP does not explicitly learn anything about the 3D point position and FK, additionally, small errors in the root or in other crucial joints influence the error distances of points in 3D, because errors are propagating through the FK computational chain.
% Moreover, the ground truth angles are often not available in datasets.

FK predicts poses in the camera frame, therefore, the loss in 3D is the Euclidean distance against the ground truth pose in the camera frame $||\hat{\mathbf{X}}_C - \mathbf{X}^*_C||$.
% The disadvantage of using 3D loss for human pose stems from the fact that the body has a higher contribution in terms of errors and weights in MLP training than does the hands.
% This problem can be mitigated by using higher weights for hands.
% However, in the experiments (see~\ref{section:experiments}) two separate models for hands and body are used.
In the experiments, two separate models for hands and body are used to avoid the fact that body pose errors are significantly higher than errors on hands, which can be problematic for training.
This is useful as it allows the networks to be applied independently when only body or hands points are available.

A slightly worse performance on the validation set comes from the ground truth 2D points supervision when the reprojection error function is used ({\ie} $||\hat{\mathbf{X}}_I$ - $\mathbf{X}^*_I||$), because a small change in estimated 3D pose results in a less accurate 2D projection with respect to the ground truth 2D points.
Therefore, the network has to learn more about correct projections than a 3D pose, and it also implies significantly longer training.
% Additionally, the training is significantly longer as the network has to learn the right projections.
% The worst performance in terms of error in 3D space and visual appearance shows reprojection error minimization between projected poses $\hat{\mathbf{X}}_C$ by $\mathbf{K}$ and image points $\mathbf{X}^*_I$.
% Here the problem is that completely different 3D poses can have the same projection in 2D, hence a low reprojection error does not imply a correct 3D pose.
%Assuming this, any combination of reprojection error with an angular or 3D loss does not have to improve the MLP model.
% Although, the positive side is that 2D error minimization can be used for semi-supervised learning.
% The image points can be normalized,~{\ie} to have the same scale and origin, and the intrinsic matrix can be set to an identity.
% % The origin for 3D pose in the camera frame can be assumed to be close to the normalized points.
% The origin for 3D pose in the camera frame can be estimated, so projection of 3D pose is close to the normalized points.
% Eventually, the MLP takes normalized image points, produces angles, and using FK generates 3D points, and projects the pose via identity matrix.
% The loss is computed against input normalized image points, hence no GT information is required.
%Note, that the proposed point normalization is done for every image to generalize input of the MLP and avoid the different location and scale of image poses.

%\subsection{Network training}
The primary loss function used in the experiments is the Euclidean distance versus GT pose in 3D.
% Supervision using the ground angles gives a higher pose error than computing the loss in 3D for the proposed method.
Comparable performance is obtained using a combination of angular and 3D loss functions. 
% However, it is marginally worse than pure Euclidean pose error.
The authors of~\cite{levinson20neurips,Zhou2019OnTC,Villegas_2018_CVPR} also use 3D distance for error computation, while in~\cite{pavllo2018quaternet} the angular and pose losses are employed in different scenarios.
The combination of 3D and reprojection loss with suitable weights as suggested by Kendal~{\etal}~\cite{Kendall2017GeometricLF} is the best option, because it enforces the model to learn both 3D points and their image correspondence, but it requires more exhaustive computations.
% Additional supervision for the training can also be done if multiple views are available.
% A pose reconstructed from a single 2D input can be back-projected to all corresponding images with the assumption that calibration for each camera is known.
% % The reprojection error minimization is suitable if multiple views are available during the training.
% % The input is still pose points of a single image, however, when the network lifts 2D then 3D pose can be back-projected to all given corresponding images using calibration parameters for each view.
% The reprojection loss is computed over all images as the distance between projected and ground truth 2D poses.
% The accuracy of this approach is comparable if 3D loss is used, however, in experiments, this increases the training time by at least an order of magnitude.
% Therefore, such training may be suitable if no 3D ground truth is given.

In summary, the human pose reconstruction method starts from the detection of 2D points from a single image using a CNN (MediaPipe in our experiments).
Afterward, it exploits four sub-networks: two to predict joint angles and a further two for bone lengths of the body and hands.
% In summary, the human pose reconstruction method exploits 4 sub networks: two to predict joint angles and a further two for bone lengths of the body and hands.
% The proposed approach does not intend to predict the origin of a pose in 3D, since given a single image, points and scale ambiguity it is a very challenging task.
The proposed approach does not intend to predict the origin of a pose in 3D, because given only a single view this is a very challenging task, and it is not important for sign language.

The angular network and the MLP for bone lengths are trained separately to avoid the effect of one model on another.
For the angular MLP training, the ground truth bone lengths and origin are used to generate a 3D pose.
For the bone length MLP, if the GT angles are available, then the network also performs FK and calculates the error in 3D, otherwise, the loss is computed as the absolute difference to the ground truth bone values.

% If body and hand networks are used jointly, then the hand points express additional weights on the position of wrist joints in 3D space.
Since models are trained separately, a body is concatenated with hands afterward.
It is important to add additional weights to the wrist points while computing the loss in 3D, otherwise, the error on the wrist joint implies a higher error on hand joints.
The benefit of using a separate hand and body model is that only one MLP for both hands can be employed by simply flipping the input and output for the left hand.
% Without loss of generality, the MLP predicts the angles for a right hand given 2D key-points.
% The image points of a left hand can easily be transformed to the right hand by multiplying the $x$-coordinate by -1.
% Afterwards, angles of the right hand predicted by the MLP are converted to the left by switching the sign of roll and pitch angles.

%%%%%%%%%%%%%%%%%%%%%%%%%%%%%%%%%%%%%%%%%%%%%%%%%%%%%%%%%%%%%%%%
% \vspace{-0.4cm}
\section{Experiments}
% \vspace{-0.4cm}
\label{section:experiments}
For sign language, we train our model on the~\textsc{Smile} sign language dataset~\cite{ebling18}. 
%The proposed method is evaluated on three datasets: the SMILE dataset~\cite{ebling-etal-2018-smile}, \textsc{Panoptic Studio}~\cite{Joo_2017_TPAMI}, and \textsc{Human3.6M}~\cite{journals/pami/IonescuPOS14}.
%One of the reasons to generate a sign dataset is to keep the same skeleton as MediaPipe in order to compare poses of the proposed MLP and MediaPipe output.
%\subsection{~\textsc{Smile} Sign language dataset}
The ground truth for the ~\textsc{Smile} sign language dataset was created by running IK with an Adam optimizer~\cite{kingma2014method} using the PyTorch library.
The optimization is performed over joint angles, bone lengths, and translations in 3D using calibrated multiview data provided by the dataset owners.
% The joint angles undergo FK to generate 3D poses that are then back-projected to the images via known calibration and the objective function used then aims to minimize the reprojection error.
% Additional constraints are applied to the available degrees of freedom of individual joints, the feasible joint limits and the preservation of bone length.
% Each sign language video has at least 3 views that are used in our reconstruction.
% %The 2D image joints of each video sequence were detected by running MediaPipe Holistic model, and then used as input to our model.
To demonstrate the superior performance of the uplift process over generic approaches, we qualitatively compare against MediaPipe on three popular sign language datasets, \textsc{Smile} dataset~\cite{ebling18} which is Swiss German Sign Language (DSGS), the PHEONIX2014 dataset~\cite{koller15:cslr} which is German Sign Language (DGS) and the BOBSL dataset~\cite{Albanie2021bobsl} which is British Sign Language (BSL). The latter datasets were not seen/used during training and demonstrate excellent generalization across different sign languages. To provide a quantitative evaluation we first look at the accuracy on the~\textsc{Smile} dataset before retaining our model on the popular 3D pose estimation datasets \textsc{Panoptic Studio}~\cite{Joo_2017_TPAMI}, and \textsc{Human3.6M}~\cite{journals/pami/IonescuPOS14} demonstrating \emph{state-of-the-art} performance.

\begin{figure}
    \centering
    % \begin{subfigure}[b]{0.99\linewidth}

    \includegraphics[width=0.21\linewidth]{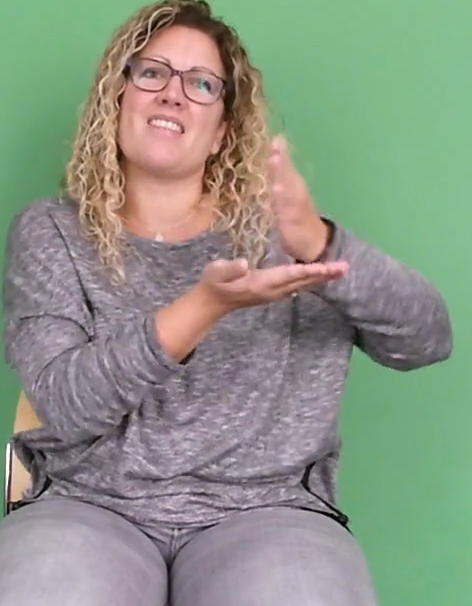}\hspace{1cm}
    \includegraphics[width=0.22\linewidth]{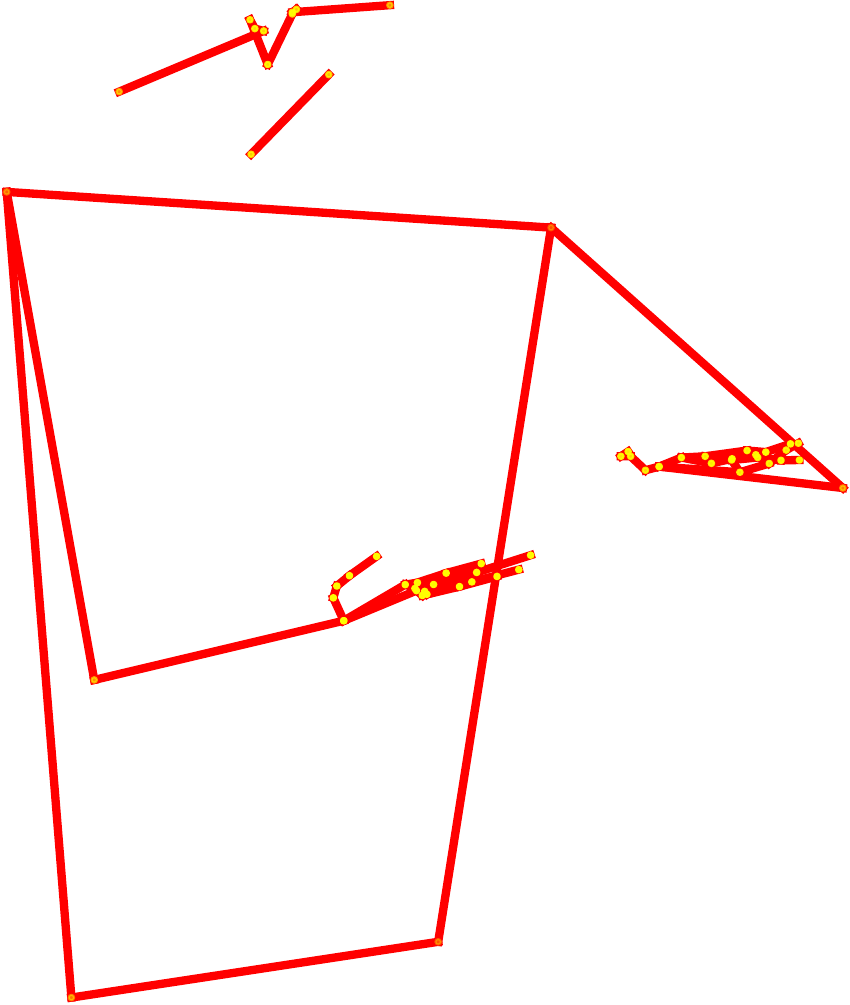}\hspace{1cm}
    \includegraphics[width=0.22\linewidth]{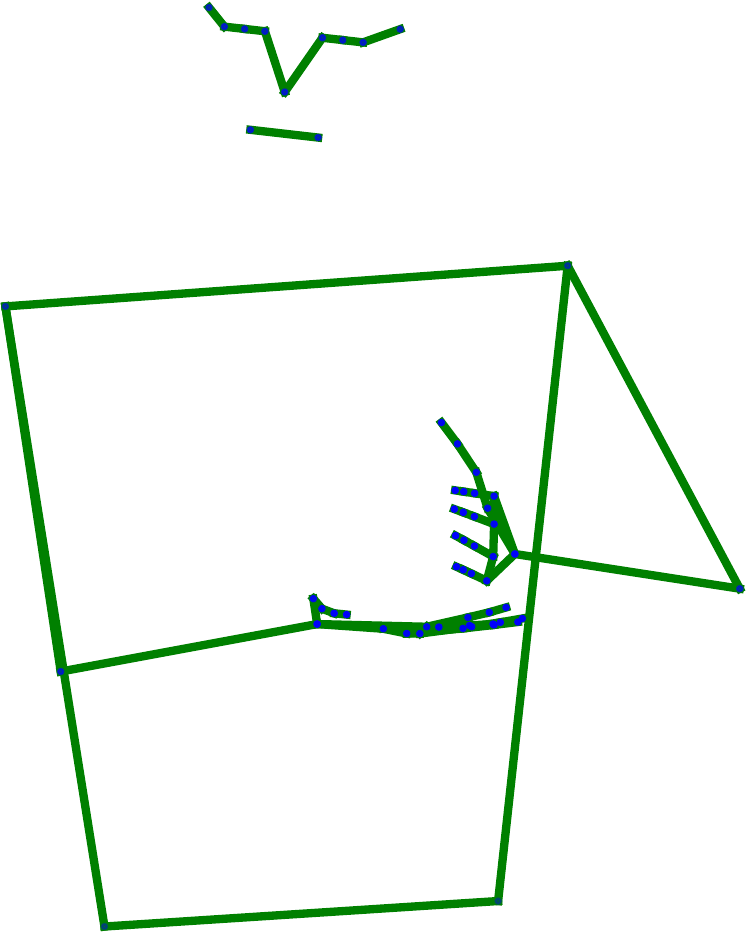}

    % \includegraphics[width=0.3\linewidth]{figures/mediapipe_comparison/0_44_bb.png}
    % \includegraphics[width=0.34\linewidth]{figures/mediapipe_comparison/0_44_mp.pdf}
    % \includegraphics[width=0.33\linewidth]{figures/mediapipe_comparison/0_44_pred.pdf}

    % \includegraphics[width=0.8\linewidth]{figures/mediapipe_comparison/0_44_bb.png}
    % \includegraphics[width=0.49\linewidth]{figures/mediapipe_comparison/0_44_mp.pdf}
    % \includegraphics[width=0.49\linewidth]{figures/mediapipe_comparison/0_44_pred.pdf}
    % \caption{$y=5/x$}
    % \end{subfigure}
    
    % \begin{subfigure}[b]{0.99\linewidth}
    % \includegraphics[width=0.8\linewidth]{figures/mediapipe_comparison/1_38_bb.png}
    % \includegraphics[width=0.49\linewidth]{figures/mediapipe_comparison/1_38_mp.pdf}
    % \includegraphics[width=0.49\linewidth]{figures/mediapipe_comparison/1_38_pred.pdf}
    % \caption{$y=5/x$}
    % \end{subfigure}

    % \begin{subfigure}[b]{0.99\linewidth}
    % \includegraphics[width=0.8\linewidth]{figures/mediapipe_comparison/7_24_bb.png}
    % \includegraphics[width=0.49\linewidth]{figures/mediapipe_comparison/7_24_mp.pdf}
    % \includegraphics[width=0.49\linewidth]{figures/mediapipe_comparison/7_24_pred.pdf}
    % \caption{$y=5/x$}
    % \end{subfigure}

%     \caption{Image in the left shows a person signing from SMILE Swiss sign language dataset~\cite{ebling18}. The pose in the center (red) is MediaPipe output, the pose in the right (green) is predicted by combination of angular and bone lengths networks. }
%     \label{fig:mediapipe_comparison}
% \end{figure}

% \begin{figure}
%     \centering

    \includegraphics[width=0.21\linewidth]{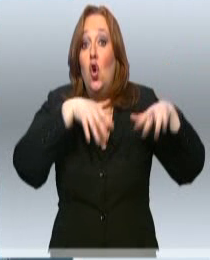}\hspace{0.85cm}
    \includegraphics[width=0.23\linewidth]{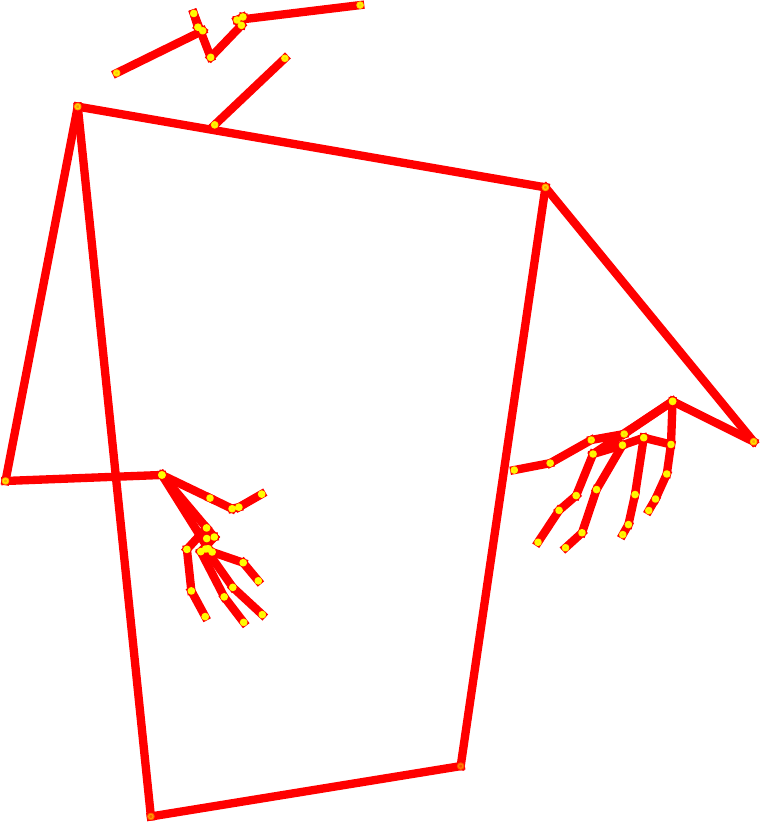}\hspace{1cm}
    \includegraphics[width=0.23\linewidth]{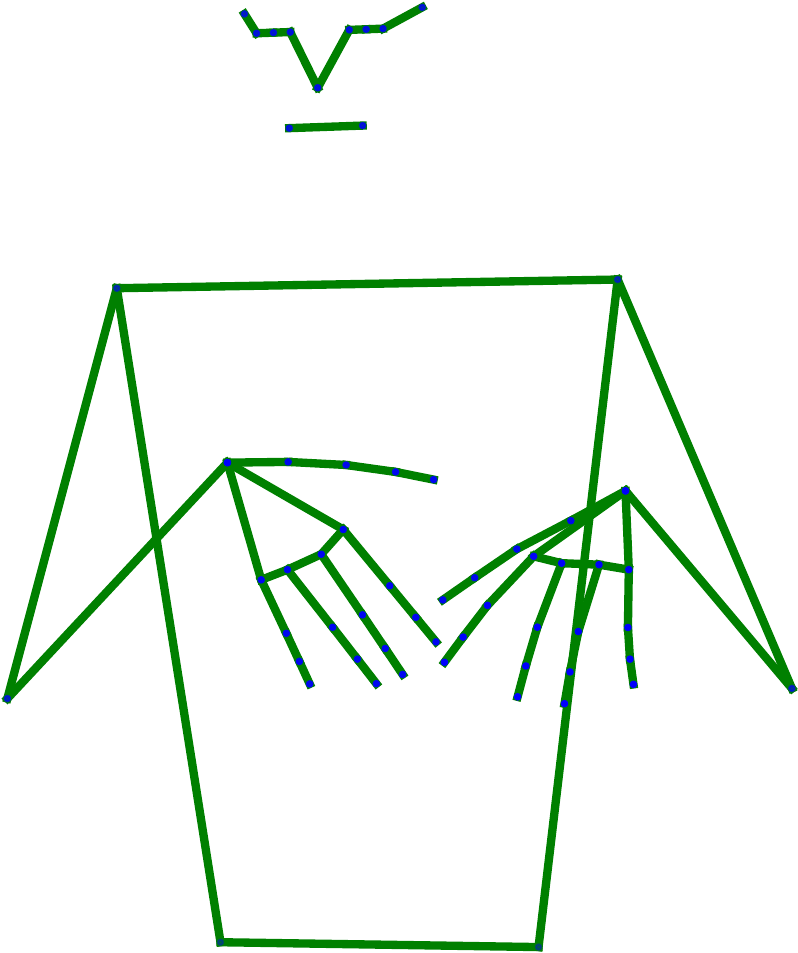}

    % \includegraphics[width=0.3\linewidth]{figures/mediapipe_comparison/1_38_bb.png}
    % \includegraphics[width=0.34\linewidth]{figures/mediapipe_comparison/1_38_mp.pdf}
    % \includegraphics[width=0.33\linewidth]{figures/mediapipe_comparison/1_38_pred.pdf}

    % \includegraphics[width=0.8\linewidth]{figures/mediapipe_comparison/1_38_bb.png}
    % \includegraphics[width=0.49\linewidth]{figures/mediapipe_comparison/1_38_mp.pdf}
    % \includegraphics[width=0.49\linewidth]{figures/mediapipe_comparison/1_38_pred.pdf}
    
%     \caption{Image in the left is from RWTH-Phoenix Weather 2014 dataset~\cite{koller15:cslr}. The MediaPipe pose is in the center (red) and proposal is in the right (green).}
%     \label{fig:mediapipe_comparison2}

% \end{figure}

% \begin{figure}
%     \centering

    \includegraphics[width=0.23\linewidth]{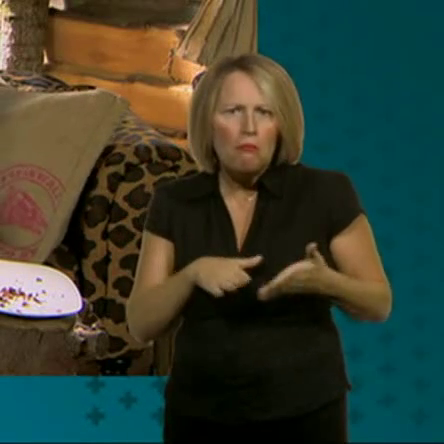}\hspace{1cm}
    \includegraphics[width=0.21\linewidth]{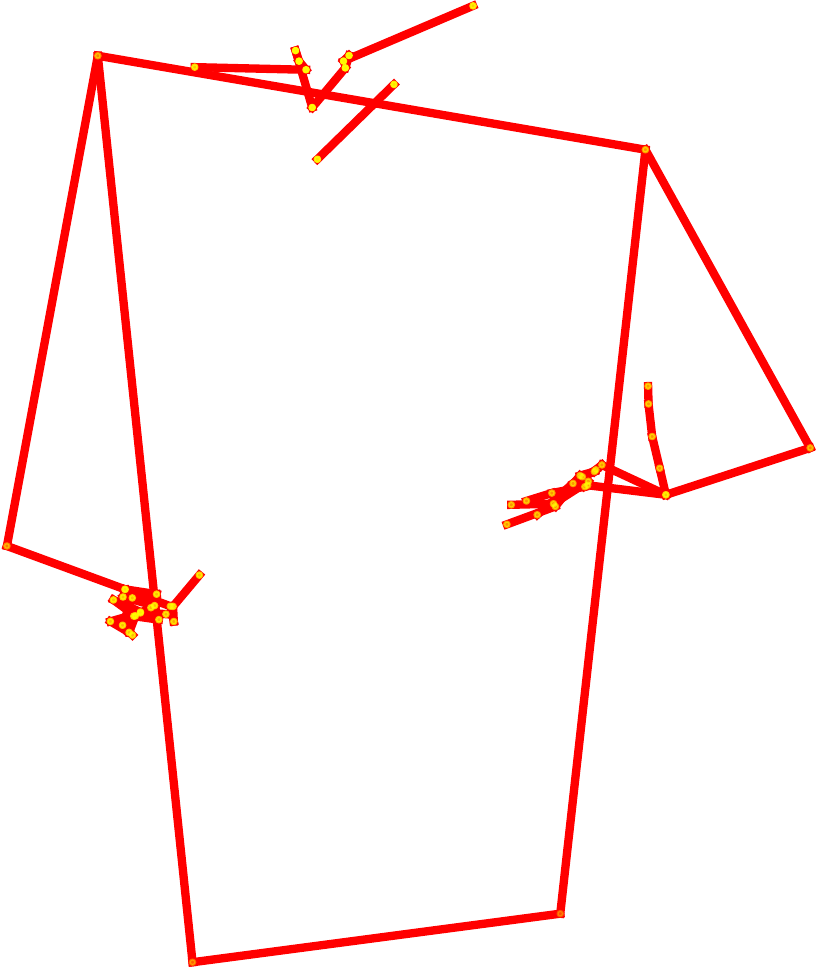}\hspace{1cm}
    \includegraphics[width=0.23\linewidth]{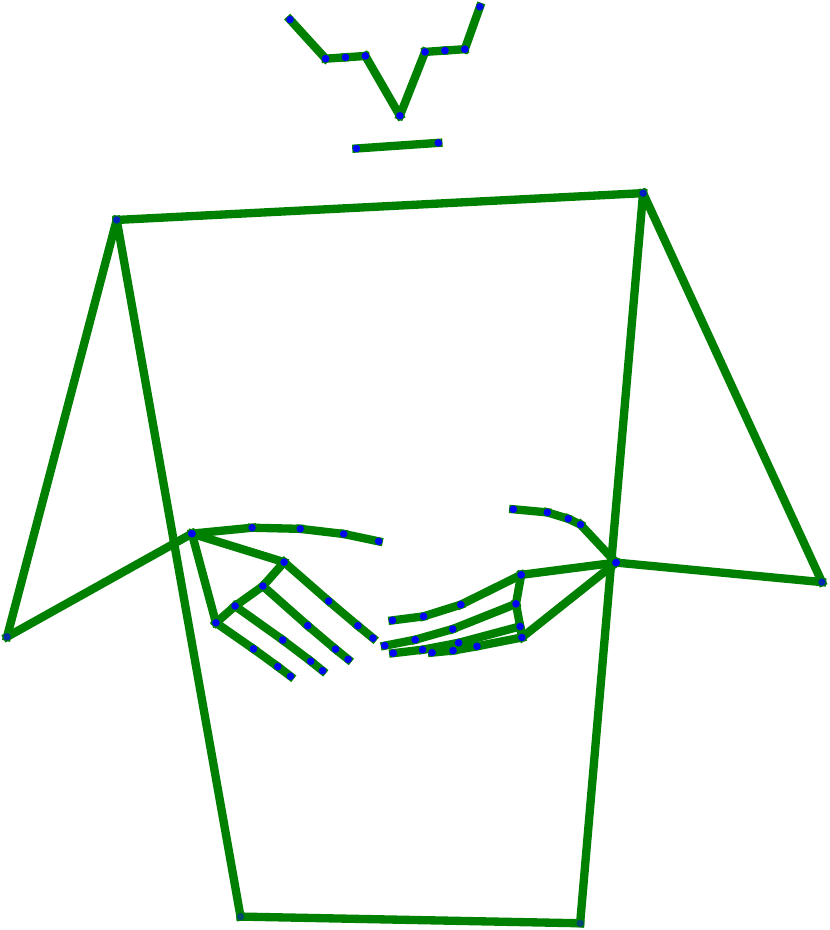}

    \caption{Images show input signer (image left), MediaPipe output (red) in the middle, and our pose (green) in the right. From the top, we show SMILE dataset~\cite{ebling18}, middle RWTH-Phoenix Weather 2014 dataset~\cite{koller15:cslr} and bottom BBC-Oxford British Sign Language dataset~\cite{Albanie2020bsl1k,Albanie2021bobsl}.
    % \vspace{-0.65cm}
    }
    \label{fig:mediapipe_comparison}
\end{figure}

% \subsection{Skeleton models}
We use MediaPipe 2D key-points of the upper torso (19 joints) as input to the lifting network since points below the waist are not relevant in the context of sign, and often not visible in an image.
The body MLP generates in total 19 angles,~{\ie} one for each elbow, 3 for each shoulder, 3 for head, 3 for the center of hips, 2 for the torso %rotation of body's part connecting the two shoulders, 
and 3 for the root, see Fig.~\ref{fig:mediapipe_comparison} for visualization.

For \textsc{Panoptic Studio} and \textsc{Human3.6M} datasets the input for the body MLP is the full MediaPipe body pose (33 points) and the target skeleton is the same as provided in datasets,~{\ie} 19 points for the whole body for \textsc{Panoptic} and 17 joints for \textsc{Human3.6M}.
The number of angles predicted by the MLP is 29 for \textsc{Panoptic} and 31 for \textsc{Human3.6M}.
A mapping of angles to joints is similar to the sign dataset with new angles added for each leg.% (hip, knee, ankle).

The angular representation of a hand is adopted from~\cite{Samadani2012MulticonstrainedIK}, {\ie} in total one hand has 26 angles. 
The \textsc{Smile} and \textsc{Panoptic} datasets have ground truth hands available, but \textsc{Human3.6M} does not.
Input for the hand MLP is again MediaPipe detection of the 21 joints of a single hand.
% Hands for \textsc{Panoptic} and MediaPipe have an equal number of points, therefore a hand MLP with the same number of parameters is used.
% The angular representation of a hand is adopted from~\cite{Samadani2012MulticonstrainedIK},~{\ie} in total one hand has 26 angles, where three angles are used for the orientation of the wrist point which is the root joint,
% eight angles are for four interphalangeal (PIP) and four distal interphalangeal (DIP) points (each has 1 DOF); one angle for thumb interphalangeal (IP), eight angles for four metacarpophalangeal (MCP) joints that have 2 DoF, and six angles for a thumb carpometacarpal (CMC) and thumb MCP (both having 3 DoF).

\subsection{Quantative evaluation}
The~\textsc{Smile} and~\textsc{Panoptic} dataset were divided into training (80\% of all video sequences), validation (10\%), and test (10\%) sets.
For the~\textsc{Human3.6M} dataset five subjects are used for training and validation while the other two for testing, this split is commonly named as protocol \#1 (see~\cite{zhaoCVPR19semantic} or~\cite{denis_lifting}).
%The best model is selected based on the validation error, while the final evaluation reported in the paper is performance on the test set, which is unseen.

% For both length we use two layers of 256 and 128 whereas the hand MLP has 3 layers with 512, 256, and 128 units. For~\textsc{Smile} the body MLP has has the same size as hand MLP but for~\textsc{Panoptic} the body MLP is $512\times256\times128\times64$, and for~\textsc{Human3.6M} the network is $4096\times512$ with dropout 30\%.%, because both datasets include complex body motion, and they need more angles to estimate.
%The best architecture was chosen with regard to performance on validation set.

% Given that the~\textsc{Smile} dataset was created by fitting 2D MediaPipe detections, both the 3D MediaPipe and the proposed method are evaluated against the ground truth reconstruction.
MediaPipe 3D was obtained by running Holistic and Hands models separately, with parameters set to non-static image, 50\% detection and tracking confidence (defaults values); and the highest model complexity available. %the model complexity for Holistic is two (highest), and the complexity for the Hand model is one (maximum).
\begin{table}[ht]
\centering
\caption{Per-joint position errors on the~\textsc{Smile} dataset. 
%MediaPipe and proposed approach are evaluated against the ground truth test poses. %Notations 
$\mathbf{R}, \mathbf{t}, s$ are rotation, translation, and scale applied to the predicted poses to align with the ground truth (GT). %Additionally, 
$\mathcal{B}$ indicates GT bone lengths were used in FK, otherwise, limb lengths are estimated by the separate network. 
Columns show median error (`median'), MediaPipe confidence weighted average error (`w. mean'), and standard deviation (`std').
%shows the weighted average error using the confidence of MediaPipe 2D detection, and `std.' column stands for the standard deviation. 
The row \#fails shows the number of times MediaPipe failed to detect hands. 
%The total number of frames used in evaluation are 802537 (37 video sequences).
}
% \begin{tabular}{|c|c|l|r|r|r|}\hline
\begin{tabular}{@{}cclrrr@{}}\hline\rowcolor{tablegray}
\multicolumn{3}{c}{} & \multicolumn{3}{c}{Per joint position error (cm)} \\\cline{4-6}\rowcolor{tablegray}
\multicolumn{3}{c}{\multirow{-2}{*}{\textsc{Smile}}} & \multicolumn{1}{c|}{median} & \multicolumn{1}{c|}{w. mean} & \multicolumn{1}{c}{std.} \\ \hline
\multirow{5}{*}{\rotatebox{90}{MediaPipe}}
& \multirow{2}{*}{\rotatebox{90}{body$\,$}}
& $\mathbf{R}, \mathbf{t}, s$ & 4.099 & 5.023& 3.099 \\
& & $\mathbf{t}, s$ & 9.155 & 11.379& 7.832 \\\cline{2-6}
& \multirow{3}{*}{\rotatebox{90}{ hand }}
& $\mathbf{R}, \mathbf{t}, s$ & 1.813 & 2.046& 1.173 \\
& & $\mathbf{t}, s$ & 3.206 & 3.768& 2.447 \\
& & \#fails & \multicolumn{3}{c}{ 443376 ($\approx 27.6\%$) } \\ 
\hline
%%%%%%%%%%%%%%%%%%%%%%%%%%%%
\multirow{8}{*}{\rotatebox{90}{ Proposed }}
& \multirow{4}{*}{\rotatebox{90}{ body }}
& $\mathcal{B}, \mathbf{R}, \mathbf{t}$  & 0.928 & 1.427& 1.584 \\
& & $\mathcal{B}, \mathbf{t}$  & 1.398 & 1.965& 1.987 \\
& & $\mathbf{R}, \mathbf{t}$ & 1.126 & 1.649& 1.647 \\
& & $\mathbf{t}$ &  1.556 & 2.134& 2.040 \\\cline{2-6}
& \multirow{4}{*}{\rotatebox{90}{ hand }}
& $\mathcal{B}, \mathbf{R}, \mathbf{t}$  & 0.573 & 0.770& 0.673 \\
& & $\mathcal{B}, \mathbf{t}$  & 0.833 & 1.143& 1.044 \\
& & $\mathbf{R}, \mathbf{t}$ & 0.629 & 0.819& 0.671 \\
& & $\mathbf{t}$ & 0.863 & 1.167& 1.034 \\
\hline
\end{tabular}
% \vspace{-0.25cm}
\label{table:sign_dataset}
\end{table}

% \begin{table}[ht]
% \centering
% \caption{Per joint position errors of test poses on \textsc{Panoptic} dataset. The column `w. mean' indicates weighted average, where weights are provided in the dataset. 
% %Additionally, hand points have 0 / 1 weighting depending on whether they are visible in the input image. The total number of frames used in evaluation is 549293 (116 video sequences).
% }
% \begin{tabular}{@{}cclrrr@{}}\hline\rowcolor{tablegray}
% \multicolumn{3}{c}{} & \multicolumn{3}{c}{Per joint position error (cm)} \\\cline{4-6}\rowcolor{tablegray}
% \multicolumn{3}{c}{\multirow{-2}{*}{\textsc{Panoptic}}} & \multicolumn{1}{c|}{median} & \multicolumn{1}{c|}{w. mean} & \multicolumn{1}{c}{std.} \\ \hline
% \multirow{8}{*}{\rotatebox{90}{ Proposed }}
% & \multirow{4}{*}{\rotatebox{90}{ body }}
% & $\mathcal{B}, \mathbf{R}, \mathbf{t}$  & 1.089 & 3.496& 5.445 \\
% & & $\mathcal{B}, \mathbf{t}$  & 1.421 & 4.378& 5.353 \\
% & & $\mathbf{R}, \mathbf{t}$ &1.471 & 4.290& 5.553 \\
% & & $\mathbf{t}$ & 1.777 & 5.067& 5.425 \\
% \cline{2-6}
% & \multirow{4}{*}{\rotatebox{90}{ hand }}
% & $\mathcal{B}, \mathbf{R}, \mathbf{t}$  & 0.558 & 1.012& 1.402 \\
% & & $\mathcal{B}, \mathbf{t}$  & 0.844 & 1.597& 1.888 \\
% & & $\mathbf{R}, \mathbf{t}$ & 0.593 & 1.052& 1.399 \\
% & & $\mathbf{t}$ & 0.874 & 1.635& 1.889 \\
% \hline
% \end{tabular}
% \label{table:panoptic}
% \end{table}

\begin{table}[ht]
\centering
\caption{Average, median, and standard deviation per-joint position errors (cm) on \textsc{Panoptic} and test subjects (S9 and S11) for~\textsc{Human3.6M} datasets.
Individual rows show a specific alignment.
%Additionally, hand points have 0 / 1 weighting depending on whether they are visible in the input image. The total number of frames used in evaluation is 549293 (116 video sequences).
}
\begin{tabular}{@{}cclrrr@{}}\hline\rowcolor{tablegray}
\multicolumn{3}{c}{} & \multicolumn{3}{c}{Per joint position error (cm)} \\\cline{4-6}\rowcolor{tablegray}
\multicolumn{3}{c}{\multirow{-2}{*}{Dataset}} & \multicolumn{1}{c|}{median} & \multicolumn{1}{c|}{w. mean} & \multicolumn{1}{c}{std.} \\ \hline
\multirow{8}{*}{\rotatebox{90}{  \textsc{Panoptic} }}
& \multirow{4}{*}{\rotatebox{90}{ body }}
& $\mathcal{B}, \mathbf{R}, \mathbf{t}$  & 1.089 & 3.496& 5.445 \\
& & $\mathcal{B}, \mathbf{t}$  & 1.421 & 4.378& 5.353 \\
& & $\mathbf{R}, \mathbf{t}$ &1.471 & 4.290& 5.553 \\
& & $\mathbf{t}$ & 1.777 & 5.067& 5.425 \\
\cline{2-6}
& \multirow{4}{*}{\rotatebox{90}{ hand }}
& $\mathcal{B}, \mathbf{R}, \mathbf{t}$  & 0.558 & 1.012& 1.402 \\
& & $\mathcal{B}, \mathbf{t}$  & 0.844 & 1.597& 1.888 \\
& & $\mathbf{R}, \mathbf{t}$ & 0.593 & 1.052& 1.399 \\
& & $\mathbf{t}$ & 0.874 & 1.635& 1.889 \\
\hline
\multirow{4}{*}{\rotatebox{90}{  \textsc{H3.6M} }}
& \multirow{4}{*}{\rotatebox{90}{ body }}
& $\mathcal{B}, \mathbf{R}, \mathbf{t}_{\text{root}}$  & 4.173  & 5.395  & 5.567 \\
& & $\mathcal{B}, \mathbf{t}_{\text{root}}$  & 4.914  & 6.429  & 6.839 \\
& & $\mathbf{R}, \mathbf{t}_{\text{root}}$ &4.747  & 5.778  & 5.412 \\
& & $\mathbf{t}_{\text{root}}$ &5.410  & 6.744  & 6.630 \\
\hline
\end{tabular}
\label{table:panoptic_human}
\end{table}

% \begin{tabular}{@{}clrrr@{}}\rowcolor{tablegray}\hline%\toprule
% \multicolumn{2}{c}{} & \multicolumn{3}{c}{Per joint position error (cm)} \\\cline{3-5}\rowcolor{tablegray}
% % \multicolumn{2}{r}{} 
% \multicolumn{2}{c}{\multirow{-2}{*}{\textsc{Human3.6M}}}
% & \multicolumn{1}{c|}{median} & \multicolumn{1}{c|}{mean} & \multicolumn{1}{c}{std.} \\\hline
% % \multirow{3}{*}{\rotatebox{90}{\footnotesize{Proposal}}} & 
% \multirow{4}{*}{\rotatebox{90}{ body }}
% & $\mathcal{B}, \mathbf{R}, \mathbf{t}_{\text{root}}$  & 4.572 & 5.715& 5.714 \\
% % & 
% & $\mathcal{B}, \mathbf{t}_{\text{root}}$  & 5.326 & 6.784& 7.019 \\
% % & 
% & $\mathbf{R}, \mathbf{t}_{\text{root}}$ & 5.066 & 6.037& 5.545 \\
% & $\mathbf{t}_{\text{root}}$ & 5.758 & 7.062& 6.830 \\
% \hline
% \end{tabular}
% \label{table:human36}
% \end{table}

Results for the~\textsc{Smile} dataset are demonstrated in Table~\ref{table:sign_dataset}.
% Since MediaPipe 3D output and the ground truth skeleton have different scale and origin, two poses were aligned via the Kabsch-Umeyama algorithm~\cite{Kabsch,Umeyama} to measure the error distance,~{\ie} per joint positional error.
% Additionally, alignment is done in terms of the rotation of the whole pose because a slight difference in pose orientation implies higher error on the rest of the joints.
Both our and MediaPipe results are aligned using Kabsch-Umeyama algorithm~\cite{Kabsch,Umeyama}, but our approach does not require scale alignment as bone lengths with approximately similar scale are predicted by the bone length MLP.
Additionally, we report errors if the ground truth bone lengths are applied.
From the Table~\ref{table:sign_dataset}, it can be seen the proposed network outperforms MediaPipe with significantly lower errors even if only a translational alignment is done.

The statistical results on the~\textsc{Human3.6M} dataset, including median and standard deviation of per-joint position error, are reported in Table~\ref{table:panoptic_human} across all test sequences.
Evaluation protocol \#1 includes aligning a predicted pose to the ground truth root joint (pelvis point, see $\mathbf{t}_{\text{root}}$ in the table).
% The~\textsc{Human3.6M} is one of the most common dataset for evaluation of 3D human pose estimation.
% Moreover, it is one of the largest datasets containing millions of frames with 3D human poses, joint rotations, and calibration.
% The poses have good accuracy as they were captured with a MoCap system.
% The professional actors perform a wide range of motion from simple to more complicated ones,~{\eg} waiting or sitting down.
Accuracy on~\textsc{Human3.6M} is comparable to \emph{state-of-the-art} which varies from 5.76 (best) to 16.21 (worst)~\cite{zhaoCVPR19semantic}.

\blfootnote{This work received funding from the SNSF Sinergia project ’SMILE II’ (CRSII5 193686), the European Union’s Horizon2020 research and innovation programme under grant agreement no. 101016982 ’EASIER’. This work reflects only the authors view and the Commission is not responsible for any use that may be made of the information it contains.}

% The~\textsc{Panoptic} dataset was created by a complex and robust procedure described in detail in~\cite{Joo_2017_TPAMI} using triangulation from multiple views, dynamic programming for skeleton generation, temporal motion cues and the trajectory of the motion etc.
% However, the reconstructed poses provided in the dataset do not have consistent bone lengths; additionally, each joint has a confidence assigned to it, and there are points that have completely failed during reconstruction.
The per-joint position errors for the~\textsc{Panoptic} dataset are shown in Table~\ref{table:panoptic_human}.
We report errors separately on body and hands, with or without using the ground truth bone lengths, and the Kabsch-Umeyama alignment.
The errors on body pose are higher than for the sign language dataset, because actors in the~\textsc{Panoptic} dataset demonstrate more complex motion.
The results for hands are similar to the sign language dataset in terms of the median, but higher for the average.
The reason for this is that the two hand models were trained separately, and the sign language dataset has better and more samples for training.
Directly comparing these results against the \emph{state-of-the-art} on~\textsc{Panoptic} is challenging as papers such as~\cite{BEV} do not train on~\textsc{Panoptic} only evaluate performance on this dataset. Despite this fact, if we look at the errors of Table~\ref{table:panoptic_human} they are roughly half of those reported in~\cite{BEV}. A fairer comparison against~\cite{RochetteGuillaume2019W3PE} uses the same dataset and evaluation protocol where we reduce the error on body estimation by over 20mm.

% The evaluation on~\textsc{Human3.6M} dataset is reported in tables~\ref{table:human36_big} and~\ref{table:human36}.
% The~\textsc{Human3.6M} is one of the most common dataset for 3D human pose estimation, and it is widely used for results comparison among many publications.
% The proposed network is compared against~\emph{state-of-the-art} methods in table~\ref{table:human36_big} beating the first six implementations but concede to the recent ones.
% It stems that most of the methods that have smaller mean per joint pose error as~\cite{zhaoCVPR19semantic,elepose,Hossain_exploiting_2018,Yang20183DHP,Fang2018LearningPG},~\cite{sun_2017_compositional,martinez_2017_3dbaseline,Zhou_2017_ICCV,mehta_2017_in_wild} regress 3D points directly using only skeleton model,~{\ie} methods do not guarantee correct joints orientation, with exception to HybrIK~\cite{li2020hybrik} by Li~{\etal} where a combination of 3D pose regression and IK constraints are used.
% The presented method always generate 3D pose with valid constraints, however, it quite depends on orientation of parent joints when running FK, {\eg} rotation of shoulder point impacts on position of a wrist, hence errors accumulate on a path from the root joint to a leaf.

% The rotational Kabsch-Umeyama alignment further decreases error, and if the ground truth limb lengths are used, the error reaches the current~\emph{state-of-the-art}.

% \vspace{-0.4cm}
\subsection{Qualitative evaluation}
% \vspace{-0.2cm}
A visual comparison of the proposed model and MediaPipe~\cite{mediapipe} is shown in Fig.~\ref{fig:mediapipe_comparison} on various sign language datasets.
% Both 2D and 3D MediaPipe poses were extracted from video sequences of the SMILE Swiss sign language dataset~\cite{ebling18}, and individual frames are used for comparison.
% Predicted poses are generated by combination of 4 models: body and hands angular MLP, and body and hands MLP for bone lengths.
Visually, poses predicted by the proposed method are better, especially, with respect to the face and hands. \looseness-1
MediaPipe hands' points are noisier without the preservation of limb length and orientation constraints.
In many cases, MediaPipe hand detection completely failed for one or both hands.%\vspace{-0.2cm}

% \subsection{Future Work}
% The future work will be addressing several issues.
% The first is a pose reconstruction from a single image of a person signing using key-point detectors for body and hands.
% Second is resolving a current problem of spatial location of hands detached from body pose, see figure~\ref{fig:failure_hands_position}.
% \input{figures/proposal_problems/problems}

% TODO:
% - show some plots from MLP training
% - multiview + monocular training
% - make table best results bold
% - write about architecture and dropout in experiments
% - time w or w/o gpu

%%%%%%%%%%%%%%%%%%%%%%%%%%%%%%%%%%%%%%%%%%%%%%%%%%%%%%%%%%%%%%%%
% \vspace{-0.2cm}
\section{Conclusions}
% \vspace{-0.2cm}
This paper presented a 3D uplift approach that combines MLPs for 3D pose reconstruction from a set of 2D pose points in a single image.
The primary contribution of this work is the combination of multiple prediction networks with a forward kinematic model that is able to generate valid and accurate 3D reconstructions. With a specific application to 3D pose estimation in sign language, the quantitative and qualitative evaluation shows that the method outperforms the commonly used MediaPipe 3D pose estimator in visual and accuracy tests. Furthermore, the model is capable of providing \emph{state-of-the-art} performance on more general 3D pose estimation datasets.

%%%%%%%%%%%%%%%%%%%%%%%%%%%%%%%%%%%%%%%%%%%%

% \vfill\pagebreak

% \newpage
% \breakpage
% \newpage
% \clearpage
% {\small
% {\footnotesize
\bibliographystyle{IEEEbib}
\bibliography{egbib}
% \addbibressource{egbib}
% }

\end{document}